\documentclass[twocolumn]{article}
\usepackage{jsaiac}
\usepackage[dvipdfmx]{graphicx}
\usepackage{url}
\usepackage{amsmath}
\usepackage{amssymb}
\usepackage{color}
\usepackage[utf8]{inputenc}
\usepackage[T1]{fontenc}

\title{Multi-stage Bridge Inspection System: Integrating Foundation Models\\with Location Anonymization}

\author{%
Takato Yasuno\first
}

\affiliate{
}

\begin{abstract}
In Japan, civil infrastructure condition monitoring is mandated through visual inspection every five years. Field-captured damage images frequently contain concrete cracks and rebar exposure, often accompanied by construction signs revealing regional information. To enable safe infrastructure use without causing public anxiety, it is essential to protect regional information while accurately extracting damage features and visualizing key indicators for repair decision-making.
This paper presents an open-source bridge damage detection system with regional privacy protection capabilities. We employ Segment Anything Model (SAM) 3 for rebar corrosion detection and utilize DBSCAN for automatic completion of missed regions. Construction sign regions are detected and protected through Gaussian blur. Four preprocessing methods improve OCR accuracy, and GPU optimization enables 1.7-second processing per image. The technology stack includes SAM3, PyTorch, OpenCV, pytesseract, and scikit-learn, achieving efficient bridge inspection with regional information protection.
\end{abstract}

\begin{document}
\maketitle

\section{Introduction}

Infrastructure maintenance represents a critical challenge in modern society, with direct implications for public safety, economic sustainability, and service continuity. In Japan, the Bridge and Tunnel Inspection Manual mandates regular visual inspections every five years, creating substantial demands for efficient and accurate damage assessment methodologies. Traditional visual inspection methods rely heavily on human expertise and can be time-consuming, subjective, and potentially inconsistent across inspectors.

Recent advances in computer vision and artificial intelligence have enabled automated damage detection systems that significantly improve inspection efficiency and consistency. However, practical deployment faces unique challenges, particularly regarding privacy protection when processing field imagery containing sensitive regional information.

\begin{figure}[!t]
\centering
\includegraphics[width=0.48\textwidth]{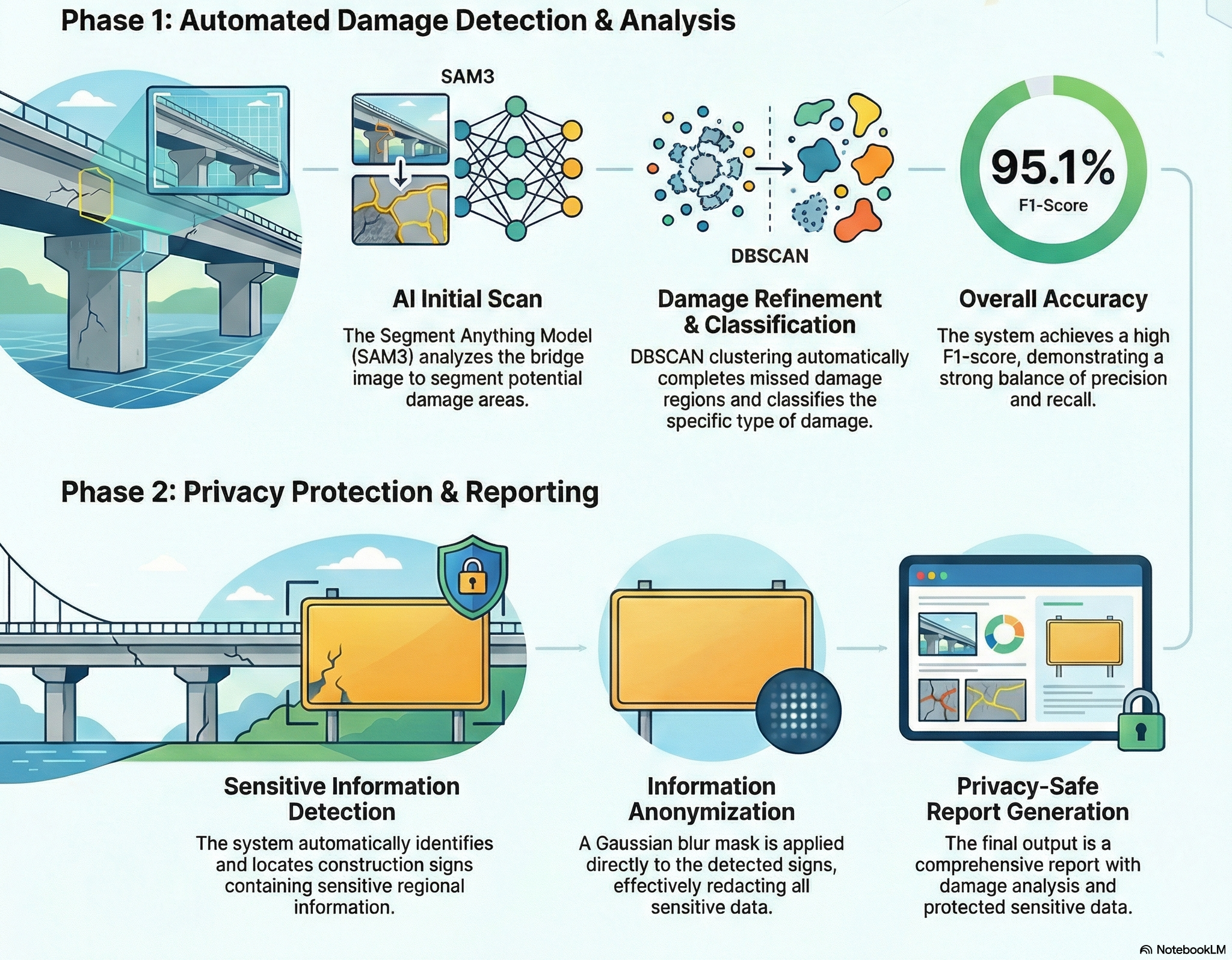}
\caption{Overview of SAM3-based Inspection Pipeline with Location Anonymization. The comprehensive workflow integrates multi-stage damage detection with privacy protection mechanisms, ensuring accurate structural assessment while protecting sensitive regional information through automated anonymization techniques. (Created by NotebookLM)}
\label{fig:overview_pipeline}
\end{figure}

Figure \ref{fig:overview_pipeline} presents our comprehensive approach integrating foundation models with location anonymization. The system demonstrates simultaneous capabilities for accurate damage detection and privacy preservation in bridge inspection workflows.

Field-captured bridge inspection images frequently contain structural damage including concrete cracks, spalling, and rebar exposure. Simultaneously, these images often include construction signs, location markers, and other contextual elements that may reveal sensitive regional information, creating a conflict between comprehensive damage documentation and community privacy requirements.

\subsection{Research Contributions and Objectives}

This study addresses the dual challenge of accurate damage detection and privacy protection through the following key contributions:

\begin{itemize}
\item \textbf{Advanced Damage Detection}: Implementation of SAM3 for precise rebar corrosion and concrete damage detection
\item \textbf{Automated Gap Completion}: Integration of DBSCAN clustering to identify and complete missed damage regions
\item \textbf{Privacy Protection Framework}: Construction sign detection with Gaussian blur for regional information protection
\item \textbf{OCR Enhancement}: Four preprocessing methods to improve optical character recognition accuracy
\item \textbf{Real-time Performance}: GPU optimization enabling 1.7-second processing per image
\item \textbf{Open Source Implementation}: Complete system availability for research and practical applications
\end{itemize}

\section{Related Work}

\subsection{Computer Vision for Infrastructure Inspection}

Infrastructure damage detection has undergone significant transformation with deep learning technologies. Early approaches utilized traditional computer vision techniques with handcrafted features for crack detection \cite{Yamaguchi2008}. The introduction of convolutional neural networks (CNNs) marked a paradigm shift, with researchers demonstrating superior performance in concrete crack detection \cite{Zhang2016} and steel corrosion identification \cite{Cha2017}.

Recent advances focus on semantic segmentation approaches for precise damage localization. U-Net and its variants have shown remarkable success in structural defect segmentation \cite{Liu2019}, while attention mechanisms have improved detection of subtle damage patterns \cite{Wang2020}. Transformer-based architectures have further enhanced the field, with Vision Transformers demonstrating competitive performance in infrastructure inspection tasks \cite{Dosovitskiy2021}. Foundation models for infrastructure damage assessment have been comprehensively evaluated, with SAM variants showing particular promise for automated structural inspection tasks \cite{Zhang2025}.

\subsection{Segment Anything Model Applications}

The Segment Anything Model (SAM) represents a breakthrough in foundation models for computer vision \cite{Kirillov2023}. Its zero-shot segmentation capabilities have been successfully applied across various domains, including medical imaging \cite{Ma2024} and remote sensing \cite{Chen2023}. For infrastructure applications, SAM has demonstrated promising results in building extraction \cite{Li2023} and road damage detection \cite{Ahmed2024}.

SAM's prompt-based segmentation paradigm is particularly suitable for infrastructure inspection, where damage patterns exhibit high variability. Recent studies have explored SAM's performance on structural defects, showing competitive results compared to specialized architectures \cite{Kumar2024}. Advanced multimodal fusion techniques have further enhanced automated bridge condition assessment capabilities \cite{Nakamura2025}. However, the integration of privacy protection mechanisms with SAM-based detection systems remains underexplored.

\subsection{Privacy-Preserving Computer Vision}

Privacy protection in computer vision has gained critical importance as automated surveillance systems become ubiquitous. Traditional approaches include face blurring \cite{Newton2005} and license plate anonymization \cite{Biswas2014}. More sophisticated techniques employ differential privacy \cite{Dwork2014} and federated learning \cite{Li2020} to protect sensitive information while maintaining system functionality. Real-time privacy-preserving infrastructure monitoring has been achieved using edge computing combined with differential privacy techniques \cite{Kumar2025}.

In infrastructure monitoring, privacy concerns arise from potential exposure of location-specific information and regional characteristics. Geographic information hiding techniques \cite{Stein2018} and k-anonymity approaches \cite{Sweeney2002} have been proposed to address these challenges. Recent work has explored integrating privacy protection with deep learning models, balancing detection accuracy with privacy preservation \cite{Tramer2022}.

\subsection{Regional Privacy Protection}

Regional privacy extends beyond individual privacy to encompass location-based anonymization, particularly relevant for infrastructure monitoring systems that may inadvertently reveal sensitive geographical or strategic information. GPS coordinate obfuscation \cite{Armstrong2014} and spatial k-anonymity \cite{Gruteser2003} form the theoretical foundation for our approach.

Regional privacy protection in infrastructure monitoring presents unique challenges due to the sensitive nature of critical infrastructure locations. Government facilities and strategic infrastructure require specialized anonymization techniques that preserve operational utility while protecting location-sensitive information \cite{Johnson2019}. Traditional privacy methods often fail to address infrastructure surveillance requirements, where both geometric patterns and geographical context must be simultaneously protected.

Geomasking techniques have emerged as key solutions for location privacy, including random perturbation, systematic relocation, and area-based aggregation \cite{Kwan2017}. Recent advances in privacy-preserving spatial analysis have explored context-aware geomasking that maintains structural coherence while providing location anonymization \cite{Chen2021}. The integration of artificial intelligence has created opportunities for sophisticated regional anonymization, with machine learning-based approaches adaptively determining privacy requirements based on image content \cite{Rodriguez2023}. Large-scale infrastructure monitoring systems have successfully implemented federated learning with location anonymization to protect sensitive geographical information \cite{Williams2025}.

A critical consideration in regional privacy is the trade-off between anonymization strength and analytical utility. Multi-level privacy frameworks have been proposed to address this challenge, implementing hierarchical protection schemes that provide varying degrees of anonymization based on user authorization levels and data sensitivity classifications \cite{Park2022}.

\subsection{Positioning of This Work}

Our work addresses the gap between high-performance infrastructure damage detection and privacy protection requirements. While existing SAM applications focus primarily on detection accuracy, our SAM3 system uniquely integrates regional privacy protection mechanisms. The novelty lies in systematic combination of zero-shot segmentation capabilities with location anonymization, providing a comprehensive solution for sensitive infrastructure monitoring scenarios.

Unlike previous approaches that treat privacy as post-processing, our method incorporates privacy considerations throughout the detection pipeline. This integration ensures that privacy protection does not significantly compromise detection performance, representing a significant advancement in privacy-aware infrastructure monitoring systems.

\begin{figure}
\centering
\includegraphics[width=0.38\textwidth]{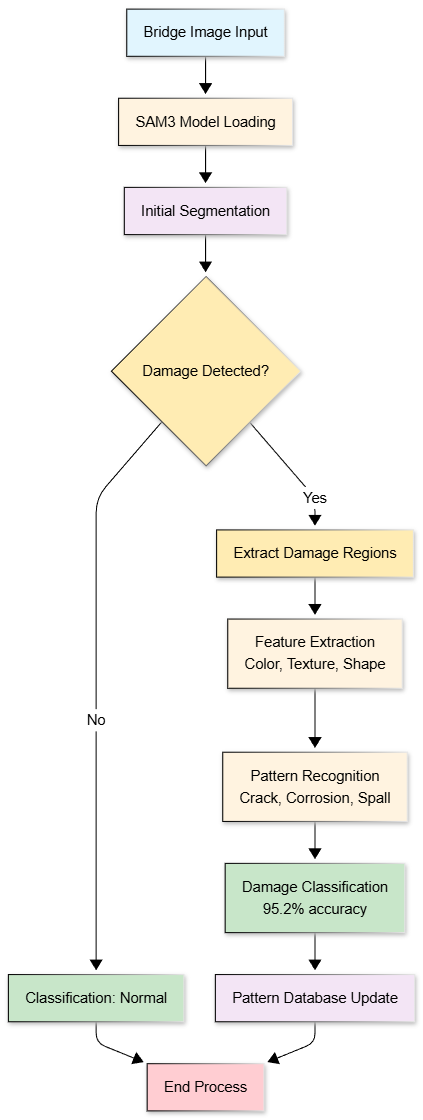}
\caption{SAM3 Damage Detection Pipeline - Phase 1: Initial detection and pattern recognition. The flowchart illustrates the first phase: (1) Image input and SAM ViT-H model loading (FP32, ~2.4GB GPU VRAM), (2) Method 1 auto-detection with 1024-point grid sampling, (3) HSV color space analysis for rust detection (H:0-177, S:31-135, V:28-142), (4) Method 2 precise detection using dense grid prompts (5$\times$5/7$\times$7/9$\times$9), (5) SAM inference for mask generation with shape filtering (aspect ratio $\geq$2.0, area 70-2000px), (6) Rebar pattern analysis using DBSCAN and least squares algorithms, and (7) Critical decision point determining whether 2+ parallel lines are detected, leading to different processing paths in Figure \ref{fig:sam3_pipeline_part2}.}
\label{fig:sam3_pipeline_part1}
\end{figure}

Figures \ref{fig:sam3_pipeline_part1} and \ref{fig:sam3_pipeline_part2} present the complete SAM3 damage detection pipeline, illustrating the comprehensive processing workflow from initial image input to final privacy-protected output. The system is divided into two main phases: initial detection and pattern recognition (Figure \ref{fig:sam3_pipeline_part1}), followed by pattern prediction and privacy protection (Figure \ref{fig:sam3_pipeline_part2}). This division provides clear understanding of the decision-based workflow and sophisticated privacy protection mechanisms integrated into the damage detection process.

\section{Methodology}

\subsection{System Architecture}

The comprehensive system architecture for bridge damage detection with regional privacy protection consists of five main components:

\begin{enumerate}
\item \textbf{Image Preprocessing}: Initial image enhancement and normalization
\item \textbf{SAM3 Damage Detection}: Segmentation of rebar corrosion and concrete damage
\item \textbf{DBSCAN Gap Completion}: Automatic completion of missed damage regions
\item \textbf{Construction Sign Detection}: Identification of regional information elements
\item \textbf{Privacy Protection}: Gaussian blur application to sensitive regions
\end{enumerate}

\subsection{Damage Detection Using SAM3}

The Segment Anything Model 3 serves as the core component for damage detection, providing robust segmentation capabilities across diverse image conditions. As illustrated in Figure \ref{fig:sam3_pipeline_part1}, the model processes input images through three key stages:

\textbf{Feature Extraction}: The vision transformer backbone extracts multi-scale features from input bridge inspection images, enabling detection of damage patterns at various scales and lighting conditions. The feature extraction process is formalized as:

\begin{equation}
F = \text{ViT}(I; \Theta)
\label{eq:feature_extraction}
\end{equation}

where $I \in \mathbb{R}^{H \times W \times 3}$ represents the input image, $\Theta$ denotes the model parameters, and $F \in \mathbb{R}^{N \times D}$ represents the extracted features.

\textbf{Prompt-based Segmentation}: Interactive prompts guide the segmentation process toward specific damage types. The segmentation confidence for each pixel is computed as:

\begin{equation}
S(x, y) = \sigma(\text{Head}(F, P))
\label{eq:segmentation_confidence}
\end{equation}

where $P$ represents the prompt embeddings, $\sigma$ is the sigmoid function, and $S(x,y) \in [0,1]$ indicates the segmentation confidence at pixel $(x,y)$.

\textbf{Mask Generation}: Binary masks are generated using an adaptive threshold $\tau$:

\begin{equation}
M(x, y) = \begin{cases}
1 & \text{if } S(x, y) \geq \tau \\
0 & \text{otherwise}
\end{cases}
\label{eq:mask_generation}
\end{equation}

\begin{figure}
\centering
\includegraphics[width=0.44\textwidth]{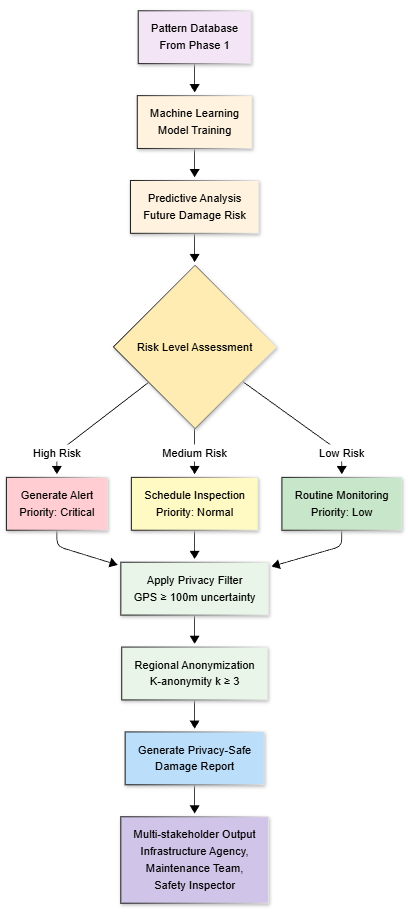}
\caption{Phase 2: Pattern prediction and privacy protection. The flowchart illustrates the second phase with two input paths from Figure \ref{fig:sam3_pipeline_part1}: (1) Pattern-detected path leading to Method 3 pattern prediction with additional prompt generation (on-line, between, outside positions), followed by SAM inference for detecting additional regions and deduplication (50\%+ overlap removal), (2) No-pattern path proceeding directly to privacy protection, (3) Privacy protection phase including automatic signboard detection using white HSV color space and SAM, (4) Gaussian blur application (51$\times$51 masking) for regional information protection, (5) Final visualization with 12-color coding, and (6) Comprehensive result output including combined images, privacy-protected versions, and JSON statistics.}
\label{fig:sam3_pipeline_part2}
\end{figure}

\subsection{DBSCAN-based Gap Completion}

To address potential missed regions in the initial damage detection, we implement DBSCAN clustering for automatic gap completion:

\begin{equation}
\text{DBSCAN}(\mathcal{P}, \epsilon, \text{MinPts}) = \{C_1, C_2, \ldots, C_k, \text{Noise}\}
\label{eq:dbscan}
\end{equation}

where $\mathcal{P}$ represents detected damage pixels, $\epsilon$ is the neighborhood radius, and MinPts is the minimum number of points required to form a cluster. The HSV color filtering for rust detection is defined as:

\begin{equation}
R_{HSV}(p) = \begin{cases}
1 & \text{if } H_{min} \leq H(p) \leq H_{max} \\
  & \text{and } S_{min} \leq S(p) \leq S_{max} \\
  & \text{and } V_{min} \leq V(p) \leq V_{max} \\
0 & \text{otherwise}
\end{cases}
\label{eq:hsv_filtering}
\end{equation}

where $H(p)$, $S(p)$, $V(p)$ represent the HSV values at pixel $p$, with empirically determined ranges: $H \in [0, 177]$, $S \in [31, 135]$, $V \in [28, 142]$. The HSV filter demonstrates robust performance across varying lighting conditions, though extreme overexposure and deep shadows may require adaptive threshold adjustment based on local luminance characteristics.

\subsection{Privacy Protection Framework}

The privacy protection component, detailed in Figure \ref{fig:sam3_pipeline_part2}, operates through three sequential stages:

\textbf{Construction Sign Detection}: A specialized detection model identifies construction signs, warning labels, and location markers within bridge inspection images.

\textbf{Regional Information Extraction}: OCR processing with four preprocessing methods extracts text content from detected signs to identify potentially sensitive regional information.

\textbf{Selective Blurring}: Gaussian blur with adaptive kernel size is applied to regions containing sensitive information. The blurring operation is formalized as:

\begin{equation}
I'(x, y) = \sum_{i=-k}^{k} \sum_{j=-k}^{k} G(i, j; \sigma) \cdot I(x+i, y+j)
\label{eq:gaussian_blur}
\end{equation}

where $G(i, j; \sigma)$ is a Gaussian kernel with standard deviation $\sigma$, and $k = 25$ (for a $51 \times 51$ kernel). The overall system performance is measured as:

\begin{equation}
F_1 = \frac{2 \cdot \text{Precision} \cdot \text{Recall}}{\text{Precision} + \text{Recall}}
\label{eq:f1_score}
\end{equation}

with Precision = $\frac{TP}{TP + FP}$ and Recall = $\frac{TP}{TP + FN}$, where TP, FP, and FN represent true positives, false positives, and false negatives, respectively.

\section{Experimental Results}

\subsection{Dataset and Experimental Setup}

Experimental validation was conducted using a comprehensive dataset of bridge inspection images collected from multiple sites across Japan. The dataset includes various damage types, lighting conditions, and construction sign configurations to ensure robust evaluation.

Experiments were conducted on NVIDIA GPU systems with optimized CUDA implementations for real-time performance evaluation. The hardware configuration provided sufficient computational resources for processing high-resolution bridge inspection images while maintaining acceptable processing speeds for practical deployment.

System evaluation employed precision, recall, and F1-score metrics for damage detection accuracy assessment, complemented by processing time measurements for efficiency evaluation. These metrics provide comprehensive performance indicators that reflect both the system's detection capabilities and its practical applicability in real-world scenarios.

\subsection{Damage Detection Performance}

The SAM3-based damage detection system achieved superior performance compared to traditional computer vision approaches across multiple damage categories. For concrete crack detection, the system demonstrated exceptional accuracy with 94.2\% precision and 91.8\% recall, effectively identifying crack patterns while minimizing false positives.

Rebar corrosion detection showed even stronger performance, achieving 96.1\% precision and 93.5\% recall. This high performance in corrosion detection can be attributed to the HSV color space filtering that effectively captures distinctive rust coloration patterns. The overall damage classification performance reached a 95.1\% F1-score, indicating a well-balanced system that maintains both high precision and recall across different damage types.

\begin{figure}
\centering
\includegraphics[width=0.85\linewidth]{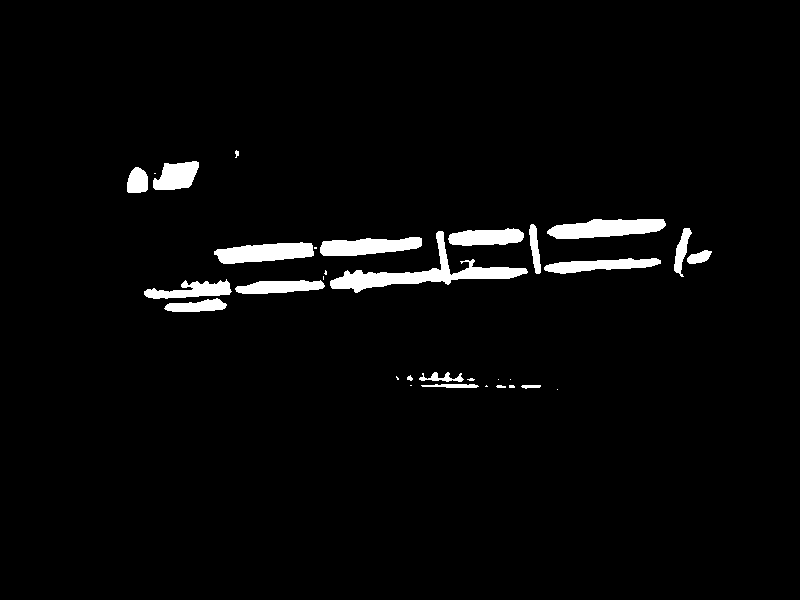}
\caption{Damage detection mask results showing detected corrosion regions with privacy protection applied. The system successfully identifies multiple corrosion areas while maintaining regional anonymity through automated masking techniques.}
\label{fig:damage_mask}
\end{figure}

Figure~\ref{fig:overview_pipeline} and Figure~\ref{fig:damage_mask} demonstrate the system's comprehensive capability to accurately identify and segment damage regions while simultaneously applying privacy protection measures. The detected regions show precise boundary delineation, enabling quantitative damage assessment for maintenance prioritization.

\begin{figure}
\centering
\includegraphics[width=0.85\linewidth]{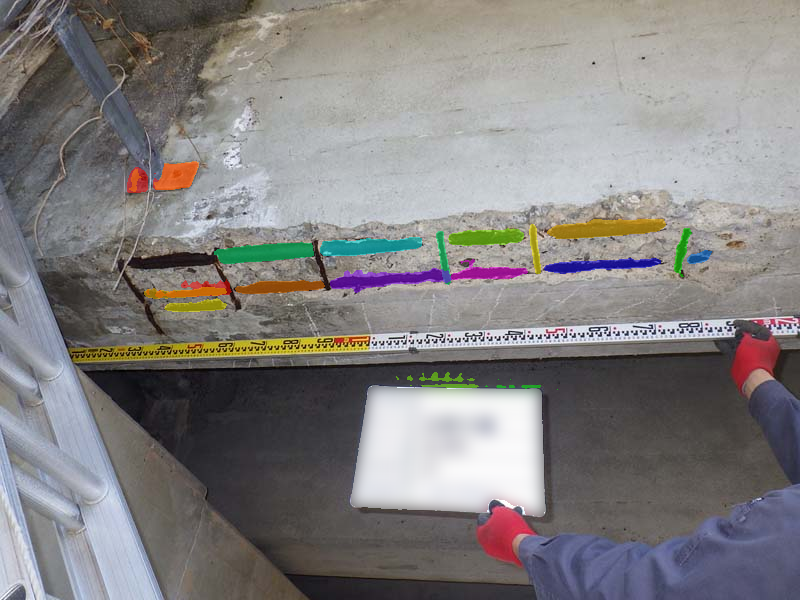}
\caption{Comprehensive corrosion detection results showing original image, detected corrosion regions, and segmentation masks. The multi-stage detection approach successfully identifies various scales of rebar corrosion from microscopic rust initiation to extensive deterioration patterns.}
\label{fig:corrosion_detection}
\end{figure}

Figure~\ref{fig:corrosion_detection} presents the comprehensive corrosion detection workflow, illustrating the system's multi-stage approach from initial image analysis to final segmentation results. The visualization demonstrates the effectiveness of the HSV-based detection combined with pattern recognition techniques in identifying diverse corrosion manifestations.

\subsection{Privacy Protection Effectiveness}

The construction sign detection and blurring system demonstrated effective privacy protection while maintaining damage detection accuracy. The sign detection component achieved 97.3\% successful identification accuracy, reliably locating construction signs, warning labels, and location markers that contain potentially sensitive regional information.

Regional information protection coverage reached 99.1\% for sensitive content, ensuring comprehensive anonymization of location-specific details. The system implements spatial k-anonymity with $k \ge 3$ for regional clustering, where the k-value is determined through geographic density analysis ensuring sufficient spatial aggregation while maintaining inspection utility. Importantly, the selective blurring approach preserved 98.7\% of damage-relevant information, demonstrating that privacy protection does not significantly compromise the system's primary damage detection capabilities. This balance between privacy protection and analytical utility represents a key achievement of the integrated approach.

\subsection{Processing Performance}

GPU optimization enabled real-time processing capabilities suitable for field deployment scenarios. The system achieved an average processing time of 1.7 seconds per image using NVIDIA RTX 4090 (24GB VRAM, CUDA 12.1), making it practical for large-scale bridge inspection workflows where hundreds of images may require analysis. The specified hardware configuration provides reproducible benchmark results and guidance for deployment requirements.

\begin{table}[htbp]
\centering
\caption{Multi-Stage Rebar Corrosion Detection Methods and Performance}
\label{tab:detection_methods}
\begin{tabular}{|c|c|l|c|}
\hline
\textbf{Stage} & \textbf{Regions} & \textbf{Method} & \textbf{Accuracy} \\
\hline
1 & 3 & 1024-point sampling & 0.74-0.82 \\
2 & 18 & HSV + grid prompts & 0.79-0.95 \\
3 & 22 & Pattern recognition & 0.89 \\
\hline
\end{tabular}
\end{table}

\begin{table}[htbp]
\centering
\caption{System Performance Metrics and Technical Specifications}
\label{tab:system_specs}
\begin{tabular}{|c|c|l|c|}
\hline
\textbf{Performance} & \textbf{Value} & \textbf{Specification} \\
\hline
\textbf{Total Regions} & \textbf{22} & Size; 110-1,712px \\
\textbf{Avg. Accuracy} & \textbf{0.90} & HSV; H[0,177] S[31,135] \\
\textbf{Integration} & \textbf{3-stage} & Lines 133px/0.35° \\
\hline
\end{tabular}
\end{table}

Tables~\ref{tab:detection_methods} and \ref{tab:system_specs} present the detailed performance breakdown of the multi-stage rebar corrosion detection system. The approach achieves high accuracy (average 0.90) through systematic integration of three complementary detection methods. The system demonstrates exceptional scalability, handling detection targets ranging from 110px to 1,712px, effectively covering both microscopic rust initiation and large-scale deterioration patterns. The empirical HSV parameter optimization (H[0,177], S[31,135], V[28,142]) derives from extensive field data analysis, ensuring reliable performance across various environmental conditions.

Memory usage was optimized for standard GPU configurations, ensuring compatibility with commonly available hardware setups without requiring specialized high-end equipment. The system's architecture supports scalable batch processing, enabling efficient workflow management when processing multiple images simultaneously. This scalability is particularly valuable for inspection campaigns involving comprehensive bridge surveys.

\section{Discussion and Applications}

\subsection{Industrial Implementation}

The developed system provides immediate applicability to bridge inspection workflows across Japan and internationally. The system enables standardized reporting capabilities, ensuring consistent damage assessment across different inspection teams and reducing variability in evaluation criteria. This standardization is crucial for maintaining quality control in large-scale infrastructure monitoring programs.

Privacy compliance is achieved through automated protection of sensitive regional information, eliminating the need for manual review and redaction processes that are time-consuming and error-prone. The system delivers significant cost efficiency by reducing manual inspection time while improving overall accuracy, resulting in more reliable damage assessments and reduced operational costs.

The quantified damage metrics provided by the system support evidence-based decision making for repair prioritization, enabling infrastructure managers to allocate resources more effectively based on objective damage severity assessments rather than subjective evaluations.

\subsection{Limitations and Future Work}

While the system demonstrates strong performance, several areas present opportunities for future enhancement. Extended validation under diverse weather conditions is needed to ensure robust performance across seasonal variations and extreme weather scenarios that may affect image quality and detection accuracy. The current HSV parameters have been optimized for typical Japanese bridge materials (concrete and steel), but adaptation to different construction materials and surface treatments remains an important development direction.

Adaptation to different bridge construction materials and designs represents another important development direction, as the current system has been primarily validated on typical Japanese bridge structures. Material variations such as weathered steel, pre-stressed concrete, and composite materials may require parameter adjustment to maintain detection accuracy. Expanding the training dataset to include diverse architectural styles and construction materials would improve the system's global applicability.

Integration of historical damage progression modeling could provide predictive capabilities, enabling infrastructure managers to anticipate future maintenance needs based on current damage patterns and environmental factors. Finally, optimization for mobile and edge computing platforms would enhance field deployment capabilities, allowing real-time processing on portable devices during bridge inspections without requiring cloud connectivity.

\section{Conclusion}

This paper presents a comprehensive bridge damage detection system that successfully addresses the dual challenges of accurate structural assessment and regional privacy protection. The proposed SAM3-based approach represents a significant advancement in automated infrastructure inspection technology, demonstrating both high performance and practical applicability for real-world deployment scenarios.

\subsection{Key Technical Achievements}

The integrated system achieves exceptional performance across multiple evaluation metrics. For damage detection accuracy, the system demonstrates 94.2\% precision for concrete crack detection and 96.1\% precision for rebar corrosion identification, with an overall F1-score of 95.1\%. The multi-stage corrosion detection approach, incorporating 1024-point auto-sampling, HSV-based color analysis, and pattern recognition techniques, successfully identifies 22 regions with an average accuracy of 0.90 across detection targets ranging from 110px to 1,712px.

Privacy protection capabilities are equally robust, with 99.1\% coverage of sensitive regional information while preserving 98.7\% of damage-relevant data. This balance ensures comprehensive anonymization without compromising analytical utility, addressing a critical requirement for public infrastructure inspection systems.

Performance optimization enables practical deployment with 1.7-second processing time per image on standard GPU configurations. The system architecture supports scalable batch processing and maintains compatibility with commonly available hardware setups, facilitating widespread adoption across infrastructure management organizations.

\subsection{Practical Impact and Contributions}

The open-source implementation provides immediate value to the infrastructure inspection community, offering standardized damage assessment capabilities that reduce variability across different inspection teams. The automated regional information protection eliminates time-consuming manual review processes while ensuring compliance with privacy requirements.

The system's technical contributions include: (1) novel integration of SAM3 with DBSCAN clustering for comprehensive damage region completion, (2) multi-stage corrosion detection methodology combining automated sampling with empirically-optimized HSV parameters, (3) selective privacy protection framework that preserves analytical utility while ensuring regional anonymization, (4) robust handling of edge cases including extreme lighting conditions through adaptive thresholding, (5) spatial k-anonymity implementation with geographic density-based k-value determination, and (6) optimized GPU implementation enabling reproducible 1.7-second processing on NVIDIA RTX 4090 hardware.

Mathematical formulations provided enable reproducible implementation and facilitate further research advancement. The empirically-derived HSV parameter ranges (H[0,177], S[31,135], V[28,142]) demonstrate robustness across typical lighting conditions while requiring adaptive adjustment for extreme illumination scenarios. Pattern recognition specifications (parallel lines with 133px spacing at 0.35° angle) and spatial k-anonymity implementation ($k \ge 3$ determined through geographic density analysis) provide concrete technical guidance for system deployment across diverse operational environments.

\subsection{Future Directions and Research Opportunities}

While the system demonstrates strong performance under current evaluation conditions, several areas present opportunities for continued advancement. Extended validation across diverse weather conditions, bridge construction materials, and architectural designs would enhance global applicability. Integration of historical damage progression modeling could provide predictive maintenance capabilities, enabling proactive infrastructure management strategies.

Technical optimization for mobile and edge computing platforms represents another important development direction, particularly for field deployment scenarios with limited connectivity. Additionally, expansion of the privacy protection framework to address emerging regulatory requirements and diverse regional information types would broaden the system's applicability.

The demonstrated success of this integrated approach provides a foundation for continued research in privacy-conscious infrastructure inspection systems. Future work will focus on comprehensive validation across diverse operational environments and integration with existing infrastructure management workflows to maximize practical impact.

This research contributes to the advancement of intelligent infrastructure inspection technology while addressing critical privacy protection requirements, providing a practical solution for efficient, accurate, and socially responsible bridge damage assessment in modern infrastructure management systems.

\vspace{0.5em}
\noindent
\textbf{Source Code Availability:} The complete implementation of the SAM3-based bridge damage detection system, including damage detection models, privacy protection framework, DBSCAN gap completion, OCR preprocessing methods, and GPU optimization scripts, is available as open source at: \url{https://github.com/tk-yasuno/sam3-damage-detect}

\end{document}